\RequirePackage{tikz}
\documentclass[sn-mathphys]{sn-jnl}

\usepackage{etoolbox}
\makeatletter
\patchcmd{\ps@headings}
{\hbox to \hsize{\hfill Springer Nature 2021 \LaTeX\ template\hfill}}
{\hbox to \hsize{\hfill Based on  Springer Nature \LaTeX\ template\hfill}}
{}
{}
\patchcmd{\ps@titlepage}
{\hbox to \hsize{\hfill Springer Nature 2021 \LaTeX\ template\hfill}}
{\hbox to \hsize{\hfill Based on  Springer Nature \LaTeX\ template\hfill}}
{}
{}
\makeatother

\theoremstyle{thmstyleone}%
\theoremstyle{thmstyletwo}%
\theoremstyle{thmstylethree}%
\usepackage{blindtext}
\usepackage{hyperref}
\usepackage{mathtools}
\usepackage{color}
\usepackage{amsmath}
\usepackage{amsfonts}
\usepackage{amssymb}
\usepackage{graphicx}
\usepackage{enumitem}
\usepackage{float}
\usepackage{longtable}
\usepackage{algcompatible}
\usepackage{algorithm}
\usepackage{caption}
\usepackage{adjustbox}
\usepackage{array}
\usepackage{rotating} 
\usepackage{booktabs}
\usepackage[justification=centering]{caption}
\usepackage[title]{appendix}
\usepackage{blindtext}
\usepackage{geometry}
\usepackage{xurl}
\usepackage{caption}
\usepackage{subcaption}

\usepackage{tikz}
\usepackage{program}
\usetikzlibrary{shapes.geometric, arrows}

\begin{document}

\title[ ]{A Machine Learning Analysis of Impact of the Covid-19 Pandemic on Alcohol Consumption Habit Changes Among Healthcare Workers in the U.S.}


\author[1]{\fnm{Mostafa} \sur{Rezapour}}\email{rezapom@wfu.edu}

\affil[1]{\orgdiv{Department of Mathematics and Statistics}, \orgname{Wake Forest University}, \orgaddress{\city{Winston-Salem}, \state{North Carolina}, \country{United States\footnote{The corresponding author current address: Biomedical Informatics, Wake Forest University School of Medicine, Winston-Salem, NC, U.S. (mrezapou@wakehealth.edu)}}}}


\abstract{In this paper, we discuss the impact of the Covid-19 pandemic on alcohol consumption habit changes among healthcare workers in the United States. We utilize multiple supervised and unsupervised machine learning methods and models such as Decision Trees, Logistic Regression, Naive Bayes classifier, k-Nearest Neighbors, Support Vector Machines, Multilayer perceptron, XGBoost, CatBoost, LightGBM, Chi-Squared Test and mutual information method on a mental health survey data obtained from the University of Michigan Inter-University Consortium for Political and Social Research to find out relationships between COVID-19 related negative effects and alcohol consumption habit changes among healthcare workers. Our findings suggest that COVID-19-related school closures, COVID-19-related work schedule changes and COVID-related news exposure may lead to an increase in alcohol use among healthcare workers in the United States.}




\maketitle

\section{Introduction}\label{sec1}

COVID-19, caused by a virus named SARS-CoV-2, was first discovered in December 2019 in Wuhan, China \cite{1}.  The Covid-19 pandemic has caused an unparalleled health crisis around the world and has brought about a negative effect on the mental health of healthcare workers \cite{7,43,44}. A rise in the rate of positive COVID-19 tests and the number of hospitalizations, reduction of proper personal protection equipment, working under severe pressures, and an increase in fears related to contracting the virus and transmitting it to others can contribute to a mental health decline among health care workers \cite{45,46,47}. Many research articles have studied the impact of the Covid-19 pandemic on the mental health decline of healthcare staff \cite{48,49,50,51,52,53}. In this paper, we apply several machine learning models and techniques to investigate relationships between COVID-19 related mental health decline of healthcare workers and their alcohol use habit changes.

Machine learning (ML) and artificial intelligence (AI) have been employed by various healthcare providers, scientists, and clinicians in medical industries in the fight against the novel disease. Researchers have utilized several advanced machine learning models and algorithms to tackle various issues related to the virus, and to better understand the pattern of viral spread. Kushwaha et al. \cite{3} discusses how ML algorithms can be used to understand the nature of the virus and predict the upcoming related issues.  Lalmuanawma et al. \cite{4} reviews the importance of AI and ML in screening, predicting, forecasting, contact tracing, and drug development for SARS-CoV-2 and its related epidemic. Benvenuto et al. \cite{5} discusses an autoregressive integrated moving average model that can predict the spread of COVID-19. Using several datasets of the COVID-19 outbreak inside and outside Wuhan, Kucharski et al. \cite{6} introduces a model that can explore the possible viral spread outside Wuhan. Machine learning has been employed to reveal the negative impacts of the Covid-19 pandemic on the students’ mental health decline \cite{54}. Machine learning methods and statistical tests have also been utilized to investigate the relationship between the COVID-19 vaccines and boosters and the total case count for the Coronavirus across multiple states in the USA as well as the relationship between several, selected underlying health conditions with COVID-19 \cite{55}. In our previous work \cite{7}, we utilized many supervised and unsupervised machine learning models to analyze the impacts the COVID-19 pandemic has had on the mental health of frontline workers in the United States.

In this study, we use survey data obtained from the University of Michigan's Inter-university Consortium for Political and Social Research (ICPSR). The data was collected by Deirdre Conroy \cite{ConroyDeirdre,conroy2021effects} from the University of Michigan Department of Psychiatry and Cathy Goldstein from University of Michigan Department of Neurology. According to the ICPSR, ``The rationale for this study was to examine whether sleep, mood, and health related behaviors might differ between healthcare workers who transitioned to conducting care from home and those who continued to report in-person to their respective hospitals or healthcare facilities." The original data contained 916 survey responses. The average survey response answered 94.5\% of the survey questions, 29 questions in total, with many questions leaving room for respondents to write how their mood or habits had changed since COVID-19 protocols were in effect. The data was stripped of any identifying information about the respondents and contained both categorical and numeric columns \cite{Web}. Rezapour and Hansen \cite{7} consider Question 29a in the survey, which reads “Please tell us how your mood has changed?” as a target variable in their analysis. They utilize multiple multiclass classification machine learning models and techniques to find the most important factors (features) in predicting the mental health decline of healthcare workers. However, this paper focuses on Question 18a: “Please tell us how the amount of alcohol that you are consuming has change?” as the target variable. In this paper, we employ many supervised and unsupervised machine learning models and techniques to analyze the hidden effects of COVID-19 on alcohol consumption changes among healthcare workers. 

The remainder of this paper is structured as follows: Section \nameref{Section2} discusses the methodology, describes the experimental framework used to find the top predictors of alcohol consumption habit changes among healthcare workers in the United States. Section \nameref{Section3} discusses and analyzes the top predictors of alcohol consumption habit changes among healthcare workers in the United States and compare the results with some other research articles. Finally, Section \nameref{Section4} concludes the paper by summarizing our overall findings.

\section{Methods}\label{Section2}
The data utilized in this study was taken from the University of Michigan’s Inter-University
Consortium for Political and Social Research, and collected by Deirdre Conroy \cite{ConroyDeirdre}. Conroy et al. \cite{conroy2021effects} confirms that all experiments were performed in accordance with relevant guidelines and regulations (see the Methods section in \cite{conroy2021effects}). The data is from a survey conducted within the University of Michigan Medical Center \cite{conroy2021effects}. All survey data was collected in accordance with relevant guidelines and regulations. The survey was undertaken after clearing University of Michigan Institutional Review Board (HUM00180147) approval, at which point the Qualtrics survey link was sent via email listservs that would reach large numbers of health care providers \cite{conroy2021effects}. It is noted that no compensation for participation was provided \cite{conroy2021effects}. Additionally, all participants have provided full and informed consent by completing the survey \cite{conroy2021effects}. 

\textbf{Data availability:} The data that support the findings of this study are available from the University of Michigan’s Inter-University Consortium for Political and Social Research, which is collected by Deirdre Conroy \cite{ConroyDeirdre,conroy2021effects}, at \url{https://www.openicpsr.org/openicpsr/project/127081/version/V1/view?path=/openicpsr/127081/fcr:versions/V1&type=project} \cite{Web}.

\textbf{Python codes availability:} Python codes for the data preparation process and other supervised and unsupervised machine learning analysis are available at \url{https://github.com/MostafaRezapour/Hidden-Effects-of-COVID-19-on-Healthcare-Workers-A-Machine-Learning-Analysis} \cite{26}. 
\subsection{Computational Process} 
     The COVID Isolation on Sleep and Health in Healthcare Workers data is a tabular dataset with 915 rows (datapoints or participants) and 64 columns (features or attributes) such as \textit{StartDate, EndDate, IPAddress}, etc. \cite{8}. The dataset contains 14678 missing values with 480 missing values for Question 28a, 311 missing values for Question 29a, and so on. To prepare the dataset for a machine learning analysis with Question 18a as the target variable, we first remove all the rows with no value (answer) in the column corresponding to Question 18a. We then remove unrelated columns such as \textit{StartDate, EndDate, Status}, etc. reducing the number of missing values to 1612.  Most of the variables in the dataset are categorical (nominal or ordinal), but since this is a COVID-related analysis, we would prefer not to use machine learning techniques or models (e.g. KNN) to treat the missing values for the categorical variables. Instead, we directly remove the rows with too many missing values from the dataset, and as the result, we end up with 273 clean rows (with no missing value). Since the mental health dataset contains categorical and ordinal variables, we first encode them to numbers, and we then use several encoding techniques such as one-hot-encoding or dummy variable encoding as well as several packages in Python such as \textit{OneHotEncoder, LabelEncoder} or \textit{OrdinalEncoder} from \textit{sklearn.preprocessing} to prepare the dataset for analysis.

In this paper, our main goal is to use multiple supervised and unsupervised machine learning models and techniques to find the top predictors (features) of alcohol consumption habit changes among healthcare workers in the U.S. during the severe phase of Covid-19 pandemic. Only for supervised machine learning models with high accuracy (at least 93\%), we discuss, export the feature importance scores and proceed feature selection phase that lead us to a reliable list of top predictors.  

We now provide a brief rerview and results for all supervised and unsupervised machine learning methods that are employed in this paper. In the supervised learning process, we train a model on a training data and evaluate its accuracy on a test data. We often split the data set into two parts, one part (for instance 75\% of all observations) as the training data set, and the other part (for instance 25\% of all data set) as the test data set. We train each model only on the training data set and then test the model on the test data set.

\subsubsection{Unsupervised Machine Learning and Feature Selection}
Here, we employ two unsupervised learning methods, Chi-squared test and mutual information, to find out the relationship between the target variable and the rest of variables. 

\noindent \textbf{Chi-Squared Test:} 

\begin{figure}[H]
  \centering
    \caption{Bar graph grouped by Question 18a and Question 11}    
  \includegraphics[width=.65\linewidth]{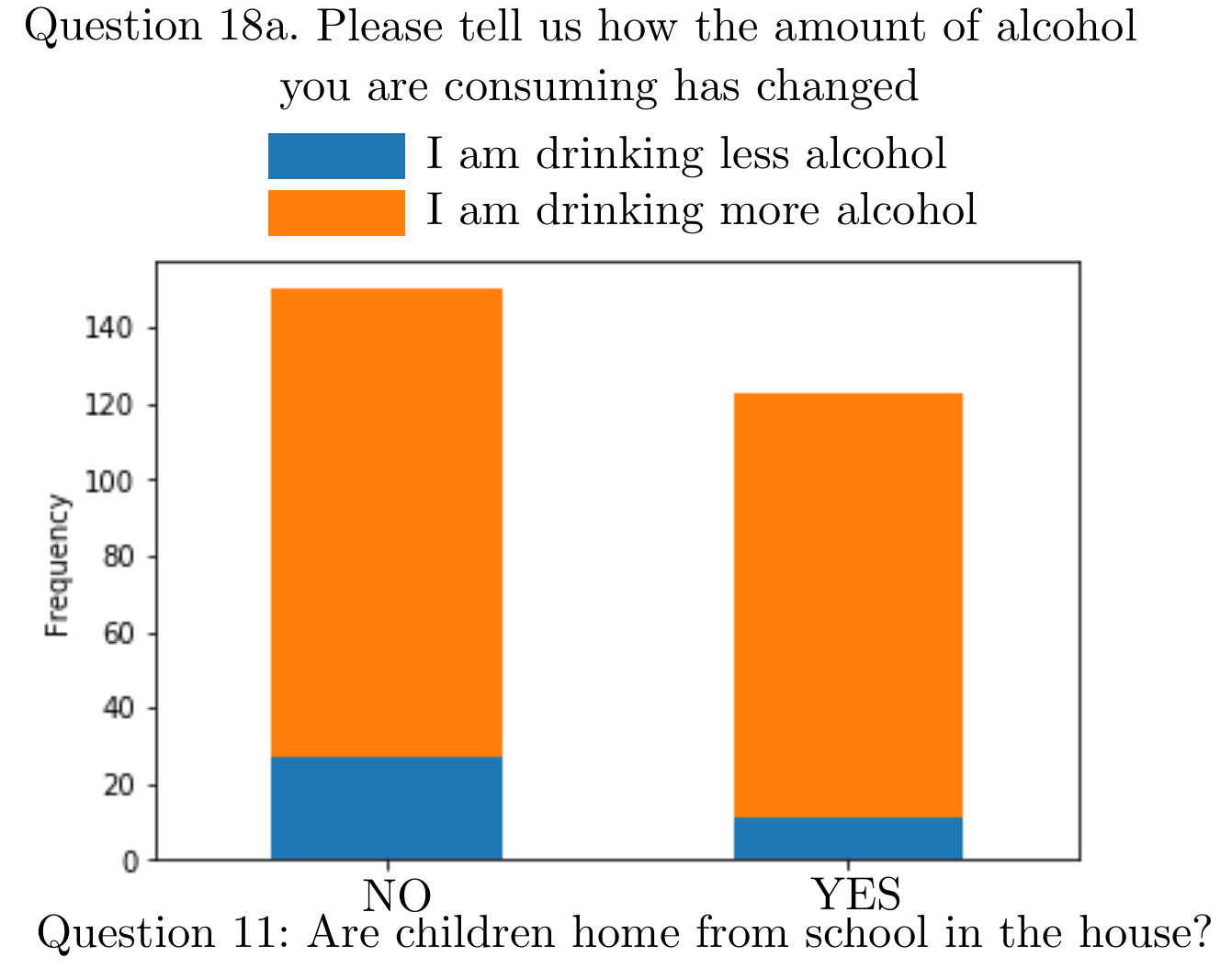}
  \label{fig2}
\end{figure}

Since the target variable and the majority of input variables are categorical (nominal or ordinal), we first apply a Chi-squared test under the null hypothesis $H_0:$ Question 18a is independent of Question $i$, where $i \in \{2, 3, 8, 9, 10, 11, 12, 13, 14, 15, 16, 17, 18, 19, 20, 21, 22, 23, 24, 25, 26, 27, 28, 29\}$ (categorical variables). Our goal is to determine whether the target variable, alcohol consumption habit changes, is independent of the input variables. By a significance level of $\alpha= 0.05$, it turns out that the null hypothesis $H_0$ is rejected for Questions 11: “Are children home from school in the house?’’ (p-value$=0.048$), Questions 15: “Have you varied your work schedule?’’ (p-value$=0.037$), and Questions 20: “In the last month, approximately how often did you have a drink containing alcohol?’’ (p-value$<0.001$). It is obvious that there is a significant association between Question 18a and Question 20. But an interesting result is the existence of sufficient statistical evidence that leads to rejection of hypothesis that Question 18a is independent of Question 11. It raises a question about the relationship between COVID-related school closures and alcohol consumption habit changes among healthcare workers. By means of bar graph, Figure \ref{fig2} displays the relationship between alcohol consumption habit changes and COVID-related school closures. One may consider parenting stress associated with COVID-related school closures as a hidden effect of the COVID-19 pandemic. 

\begin{figure}[H]
  \centering
    \caption{Bar graph grouped by Question 29 and Question 20}    
  \includegraphics[width=.65\linewidth]{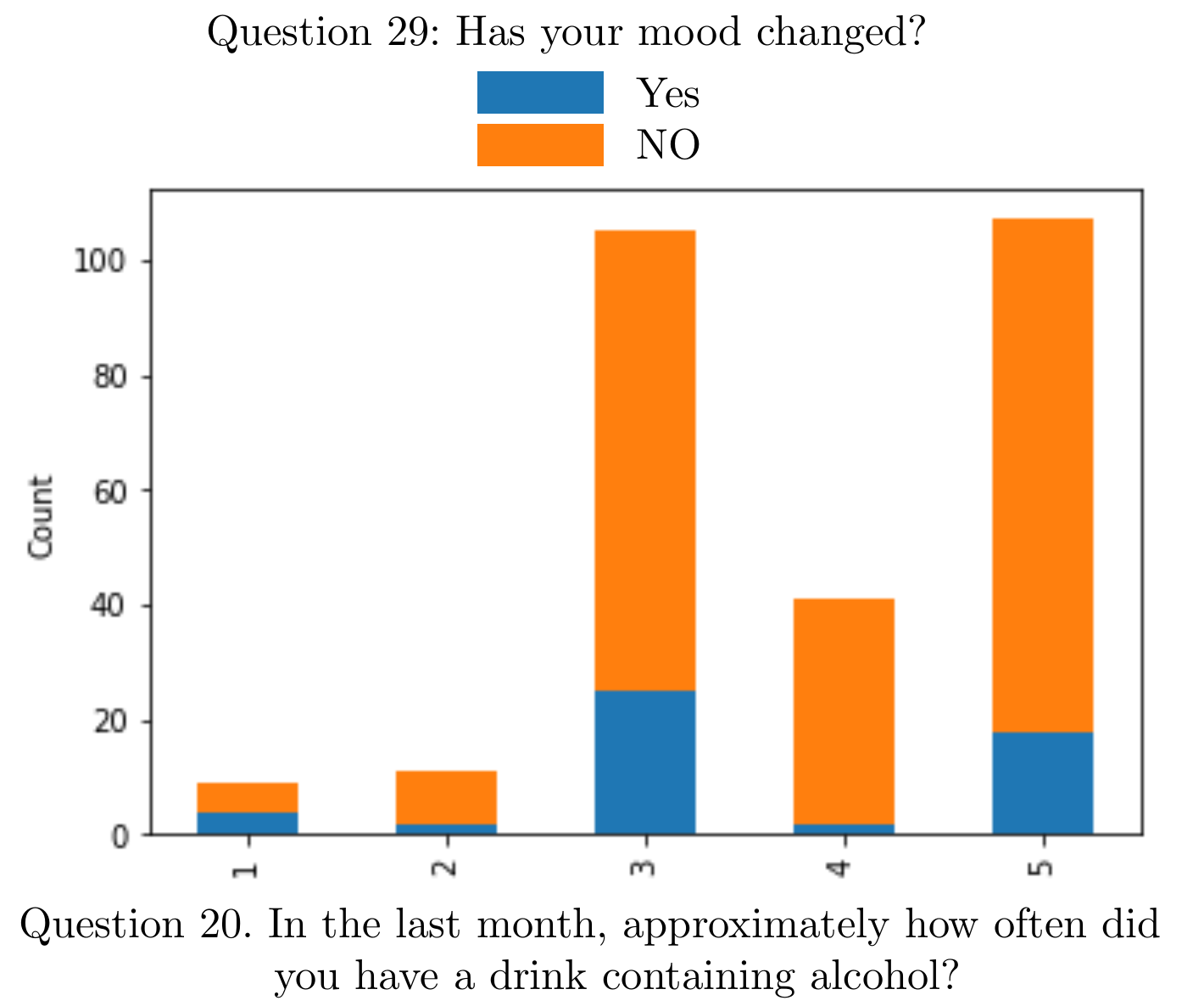}
  \label{fig3}
\end{figure}

Another question that arises is whether there is a relationship between Question 29: “Has your mood changed?’’ and Question 20: “In the last month, approximately how often did you have a drink containing alcohol?.’’ Since both variables are categorical (nominal or ordinal), we employ Chi-squared test under the null hypothesis $H_0:$ Question 20 is independent of Question 29 with a significance level of $\alpha= 0.05$ to answer the question. It turns out that the null hypothesis is rejected with a p-value equals  0.025, which indicates that there is a statistically significant association between the shifts in mood related to Covid-19 and alcohol consumption frequency (see Figure \ref{fig3}).

\noindent \textbf{Mutual Information:}

The mutual information method utilizes information gain to score features based on their importance. It can be used to measure the dependence of two variables. We use \textit{mutual-info-classif} from \textit{sklearn.feature-selection} in Python to find the most related features to Question 18a. It turns out that Question 20: “In the last month, approximately how often did you have a drink containing alcohol?,’’ Question 11: “Are children home from school in the house?’’ and Question 15: “Have you varied your work schedule?’’ obtain the highest scores among all features. Like Chi-squared test, the mutual information method also emphasizes that there exists a relationship between the parenting stress associated with COVID-related school closures and alcohol consumption changes. The parenting stress might contribute to an increase in alcohol consumption and a decline in the mental health of healthcare workers too. The findings in \cite{7}, which indicate that alcohol consumption is an important feature (factor) that may contribute to a decline the mental health of healthcare workers, agree with our Chi-squared test and Mutual Information results. 

\subsubsection{Supervised Machine Learning and Feature Selection}
We now apply supervised machine learning models and techniques to find the most important factors of alcohol consumption changes among the healthcare workers during the severe phase of the COVID-19 pandemic. Notice that Question 18a has two classes, which implies that binary classifiers must be employed for analysis. We consider an accuracy score threshold to determine whether a feature score analysis for a trained model is worthwhile or not. If accuracy score of a supervised model is above (or close to) 93\%, we analyze its feature importance scores to find the most reliable list of top predictors.

\noindent \textbf{Logistic Regression:}

\begin{figure}[H]
  \centering
    \caption{Accuracy scores of some supervised machine learning models}    
  \includegraphics[width=.65\linewidth]{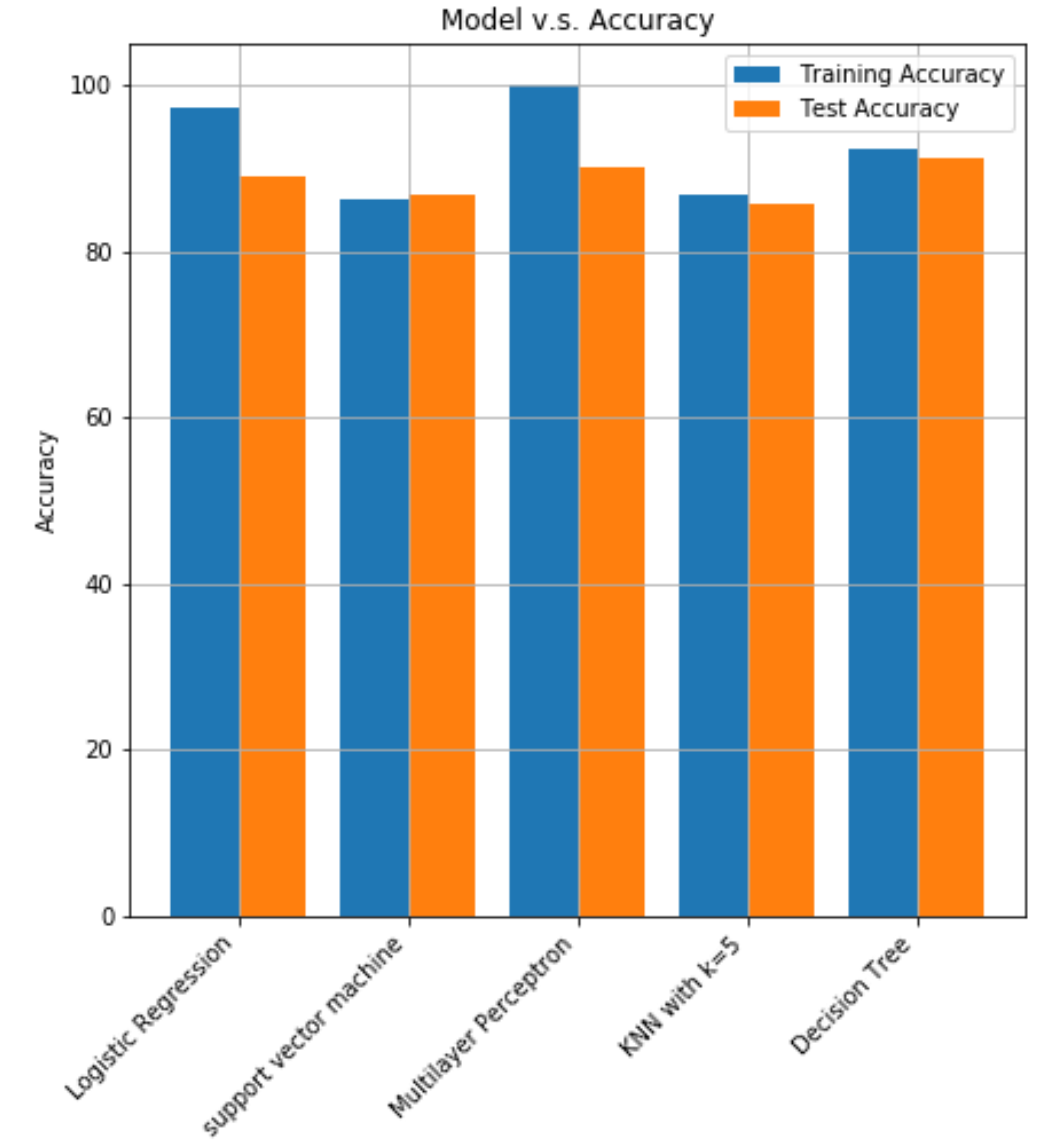}
  \label{fig4}
\end{figure}

Logistic regression \cite{9, 10} is one of the simplest supervised learning classification algorithms that can be used for a binary classification. Using logistic regression model, we can find out the relationship between the target variable and input variables because not only does it give inference about the importance of each feature, but it also gives the direction of association.  It turns out that Logistic regression accuracy on the test set is 89.01\% (see Figure \ref{fig4}), which prevents us from diving into further details about feature importance scores of this model.

\noindent \textbf{Support Vector Machines:}

Support vector machines (SVM) \cite{11, 12} is another supervised learning algorithm that can be used for both regression and classification. It is very effective in cases where the number of features is greater than the number of datapoints. SVM algorithm can be used for binary classification where it finds a separating line (or hyperplane for higher dimensions) between datapoints of two classes. A SVM algorithm finds the optimal separating hyperplane by finding the closest points (support vectors) to the hyperplane, and then maximizing the distance (margin) between the hyperplane and the support vectors. The interesting strategy that SVM algorithm utilizes for binary classification is that if the data is not linearly separable, then SVM transforms the data to a higher dimensional space, where the transformed data is linearly separable. It turns out that SVM accuracy on test set is 86.81\%, which is below our threshold for feature importance analysis (see Figure \ref{fig4}).

\noindent \textbf{Multilayer Perceptron:}

Multilayer perceptron (MLP) \cite{13, 14, 15, 16, 17, 18} is a class of feedforward artificial neural-networks that contains three types of layers (the input layer, hidden layers and output layer). In an MLP, the data flows from input to output layer, and it can be used for binary classification problems whose data is not linearly separable. One of advantages of using MLP is that it does not make any assumption regarding the underlying probability density functions. Despite MLP achieves a better accuracy on test set, 90.11\%, it does not satisfy our criteria for feature importance analysis (see Figure \ref{fig4}).  

\noindent \textbf{K-Nearest Neighbors:}

The $k$-nearest neighbors (KNN) algorithm \cite{19} is a simple supervised learning algorithm that can be used for both classification and regression. For a classification, the KNN algorithm finds the distance (e.g. Euclidean) between a query and all datapoints, and it then makes a sample space with the $k$ nearest datapoints to the query, and it finally returns the majority class as the prediction for the class of the query. The KKN algorithm is often slow, computationally expensive, and it may not work well with large datasets. Like the previous supervised models, KNN accuracy on test set, 85.71\%, is not good enough for going through further details of the results obtained by the model (see Figure \ref{fig4}).  

\noindent \textbf{Decision Tree:}

Decision trees (DT) \cite{19} are nonparametric supervised learning models that can be used for both classification and regression. A DT contains nodes (the root node, intermediate nodes, and leaf nodes) and branches.  For categorical decision trees, certain metrics, such as the Gini index or the Entropy, are used to split the training data to into smaller and smaller subsets containing data points that are more homogenous. A DT can be used for finding out the relationship between the target variable and the input variable, or feature selection. It turns out that DT accuracy on the test set is 91.21\% (see Figure \ref{fig4}), but it is not satisfactory.  

\noindent \textbf{Extreme Gradient Boosting (XGBoost):}

\begin{figure}[H]
  \centering
    \caption{Accuracy scores of XGboost (100 trees, depth=3) k-fold cross validation when k is changing between 2 and 40}    
  \includegraphics[width=.8\linewidth]{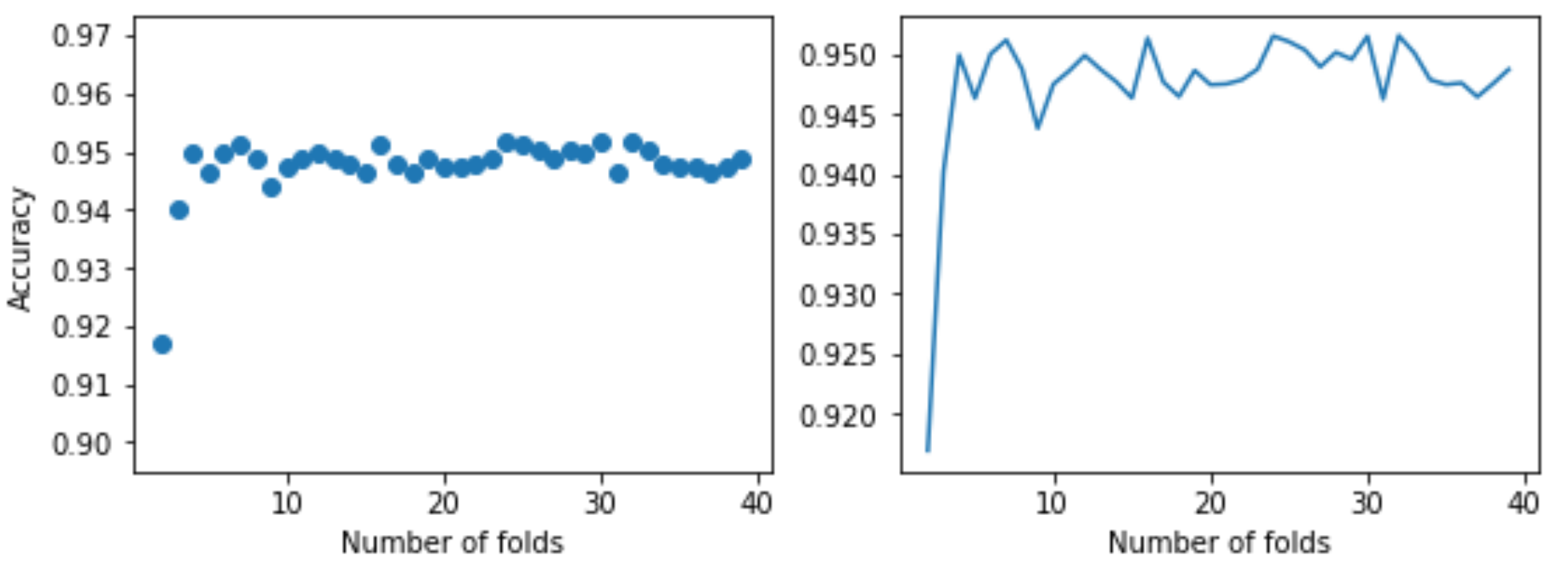}
  \label{fig5}
\end{figure}

XGBoost \cite{20, 21} is a decision-tree based ensemble supervised learning algorithm that follows the principle of gradient boosting, but it is more regularized.  XGBoost was initially introduced to improve GBM’s training time, and it combines the estimates of a set of decision trees (weaker learner) to predict the output of a target variable. In the learning process, XGBoost minimizes a regularized loss function. In the learning process, new decision-trees are added iteratively, one by one, in order to correct the prediction of the previous trees in the model. XGBoost algorithm has several hyperparameters such as number of trees, depth of each tree and number of datapoints used to train the model. 
To control number of datapoints used for training process, we combine a XGBoost and $k$-fold cross-validation \cite{42}. Figure \ref{fig5} displays accuracy scores of  a XGboost with 100 trees with depth 3, and $k$-fold cross-validation, where $k$ changes between 2 and 40.

\begin{figure}[H]
  \centering
    \caption{Feature scores of XGboost with 100 trees, depth equals 3, and $k=9$ cross-validation}    
  \includegraphics[width=.7\linewidth]{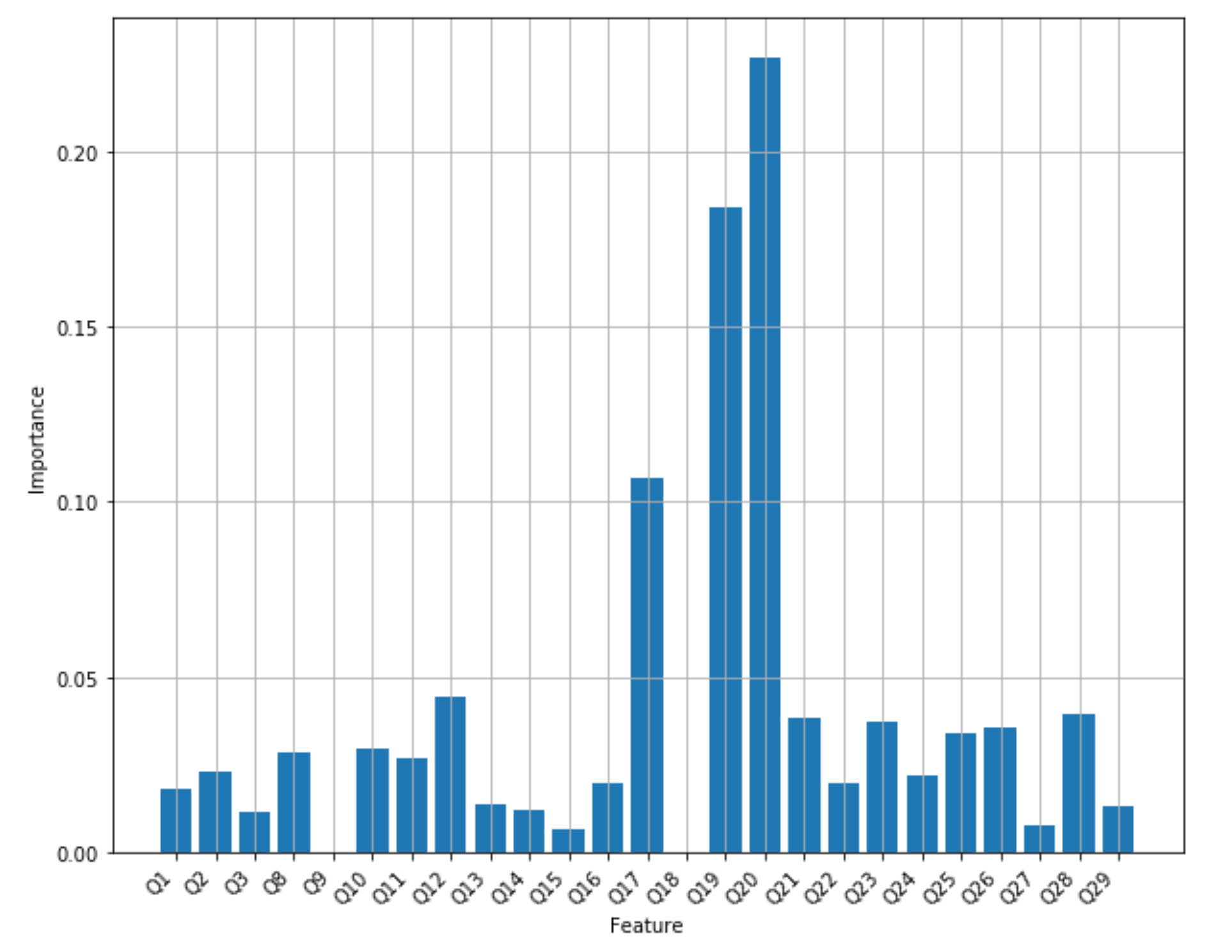}
  \label{fig6}
\end{figure}

\begin{figure}[H]
  \centering
    \caption{Q18a versus Q18 bar graph group}    
  \includegraphics[width=.7\linewidth]{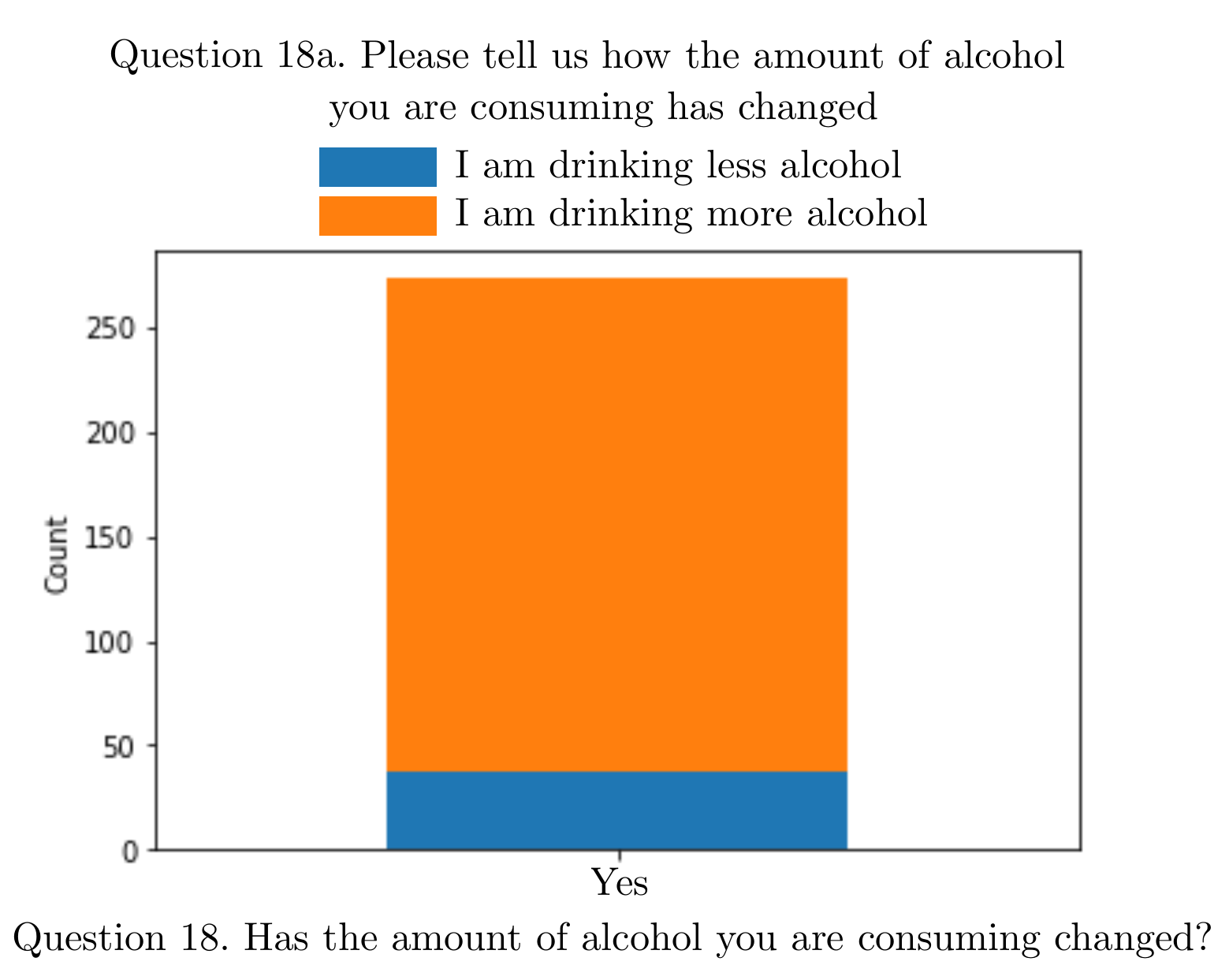}
  \label{fig7}
\end{figure}

Figure \ref{fig5} indicates that a XGboost with 100 trees with depth 3 and $k=4$ cross-validation returns an accuracy of the model is equal to 95\%.  It also reaches an accuracy of 95.2\% if $k=32$. Figure \ref{fig6} displays the feature importance scores of the model. Figure \ref{fig7} spells out why the feature importance score of Question 18 is zero, and it is because all survey respondents answers to Question 18 is ``Yes".  Table \ref{tab1} displays the top predictors, the most important features, obtained by the XGBoost model.

\begin{figure}[H]
  \centering
    \caption{Feature scores of XGboos with 100 trees, depth equals 3, k=9 cross-validation after Q19 and Q20 are removed}    
  \includegraphics[width=.8\linewidth]{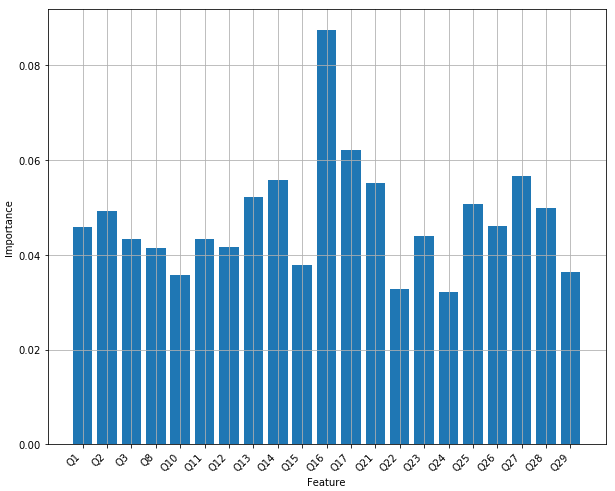}
  \label{fig8}
\end{figure}

\begin{table}
{
\begin{center}
\caption{The top predictors obtained by the XGBoost model}\label{tab1}
\scalebox{1}{%
\begin{tabular}[t]{ |l||l|  }

\hline
 \multicolumn{2}{|l|}{Question 19: “In January 2020, approximately how often} \\
\multicolumn{2}{|l|}{\hspace{1.3cm} did you have a drink containing alcohol?’’}
\\
\hline 
  \multicolumn{2}{|l|}{Question 20: “In the last month, approximately how often} \\
  \multicolumn{2}{|l|}{\hspace{1.3cm} did you have a drink containing alcohol?’’} \\
\hline 
    \multicolumn{2}{|l|}{Question 17: “Has the number of naps you are taking } \\
    \multicolumn{2}{|l|}{\hspace{1.3cm}changed?”} \\
  
  \hline
\end{tabular}
}
\end{center}
}

\end{table}
However, to get a good understanding of other features and find out the relation between alcohol consumption habit changes and other variables, we remove Questions 19 and 20, and retrain a new XGBoost on the remaining. As the result, the feature importance scores of XGBoost in absence of Questions 19 and 20 are shown in Figure \ref{fig8}.
Figure \ref{fig8} indicates that Question 16: “Have your sleep patterns changed?’’ is the most important feature of the new model. It seems that there is a relationship between the sleep pattern and alcohol consumption habit changes among healthcare workers. 

\noindent \textbf{Light Gradient Boosted Machine (LightGBM):}

\begin{figure}[H]
  \centering
    \caption{Accuracy of LightGBM for many different numbers of trees}    
  \includegraphics[width=\linewidth]{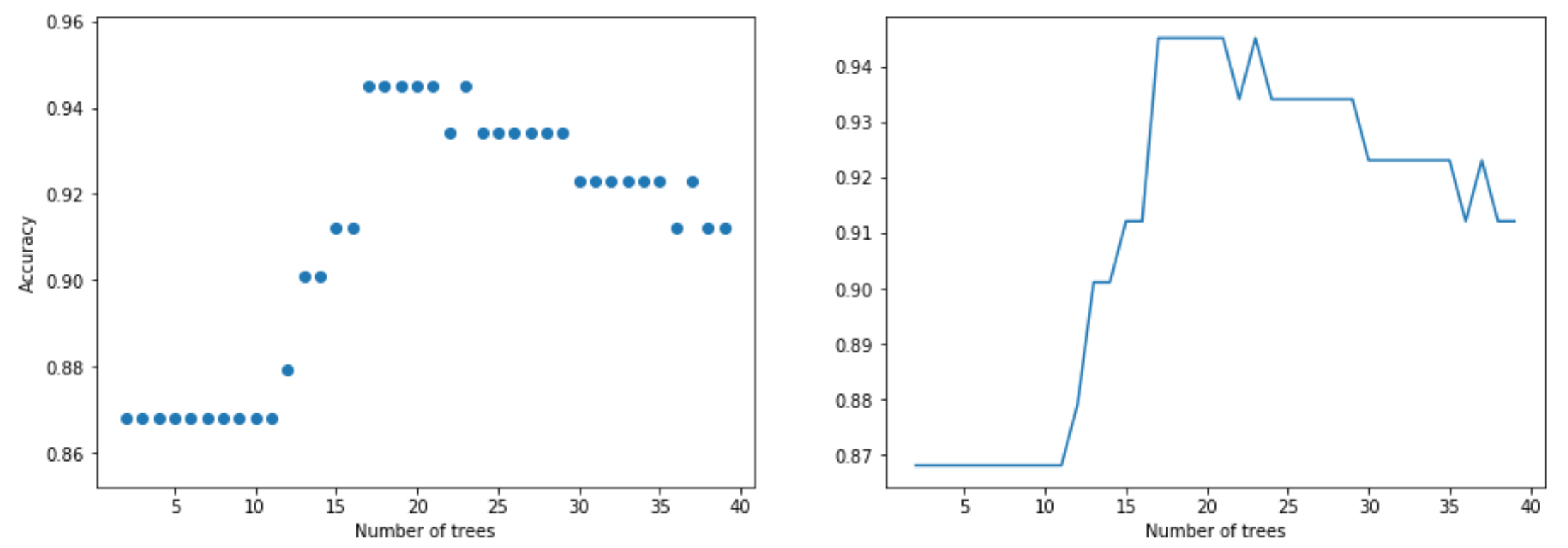}
  \label{fig9}
\end{figure}

\begin{figure}[H]
  \centering
    \caption{Feature scores of LightGBM}    
  \includegraphics[width=.8\linewidth]{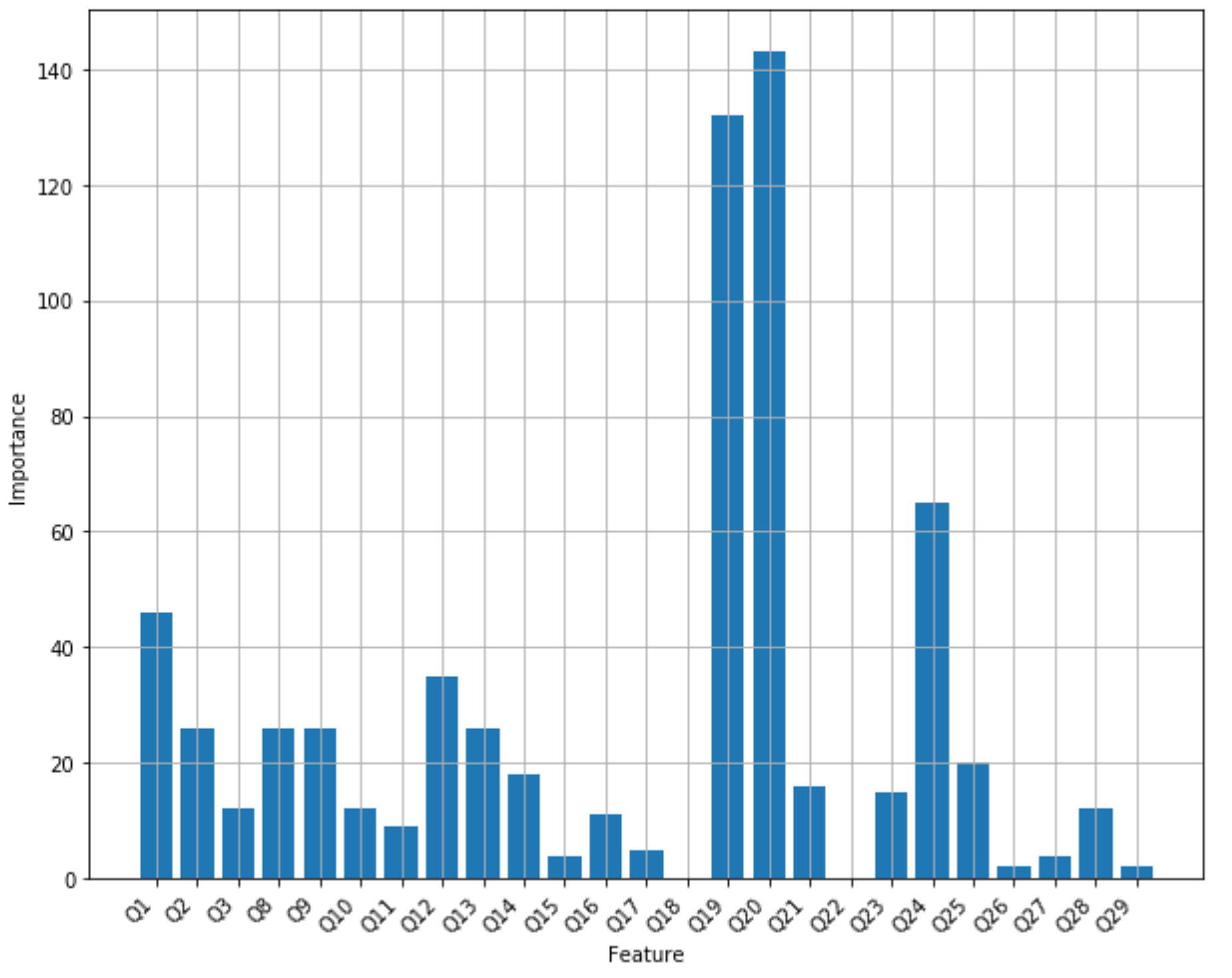}
  \label{fig10}
\end{figure}

\begin{figure}[H]
  \centering
    \caption{The accuracy scores of a 20-tree-LightGBM with different maximum depths}    
  \includegraphics[width=\linewidth]{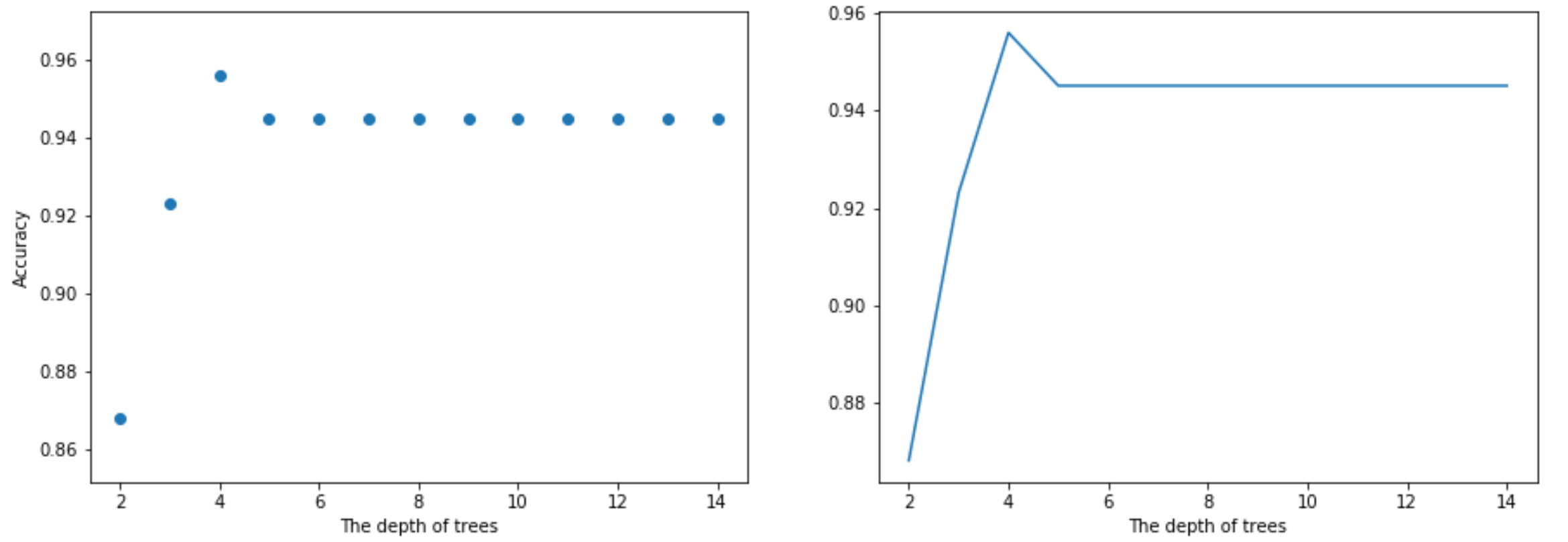}
  \label{fig11}
\end{figure}

\begin{figure}[H]
  \centering
    \caption{Feature scores of LightGBM with 20 trees (depth equals 4)}    
  \includegraphics[width=.8\linewidth]{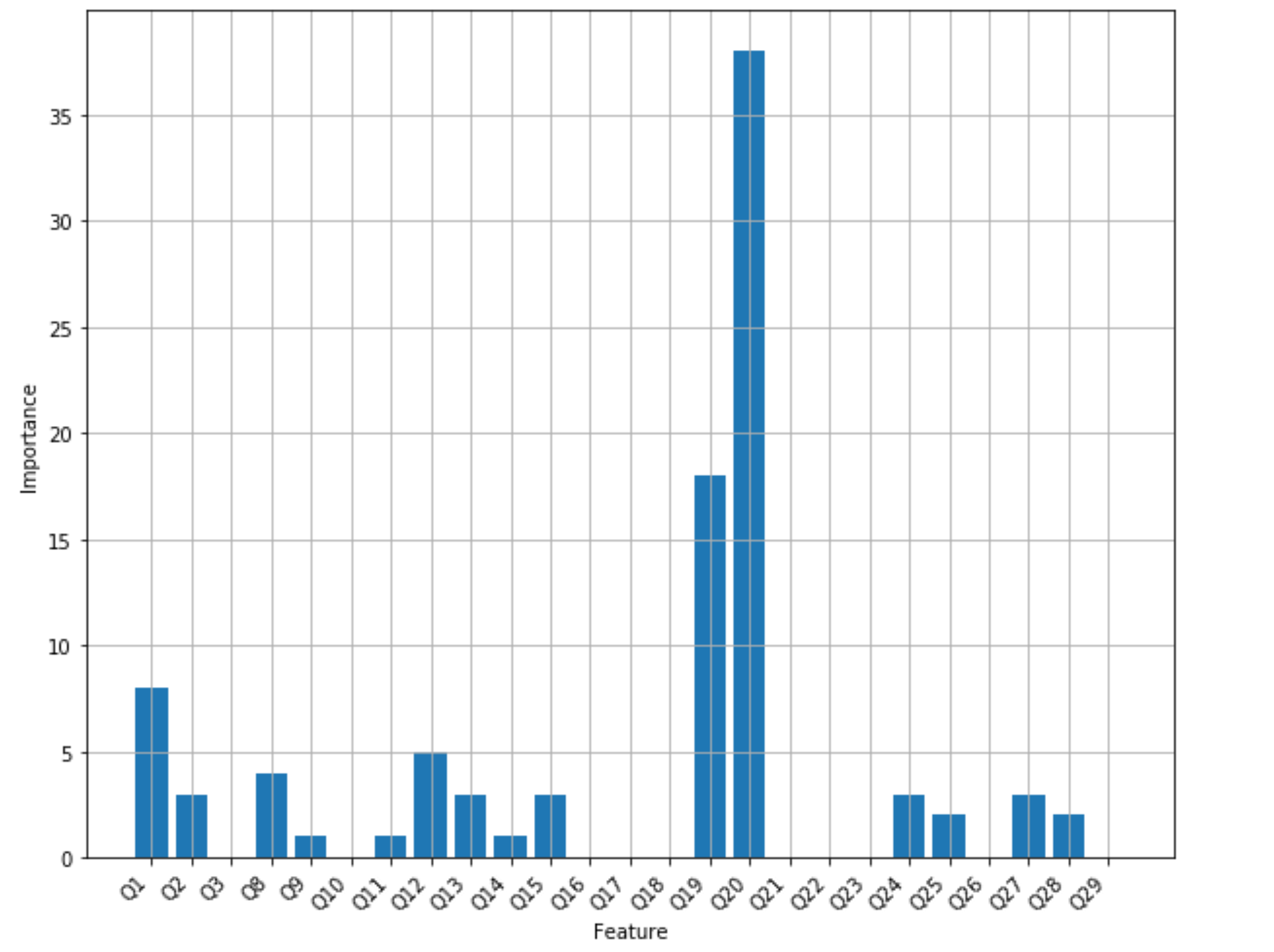}
  \label{fig12}
\end{figure}

\begin{figure}[H]
  \centering
    \caption{Feature scores of LightGBM with 20 trees (depth equals 4) in absence of Questions 19 and 20}    
  \includegraphics[width=.8\linewidth]{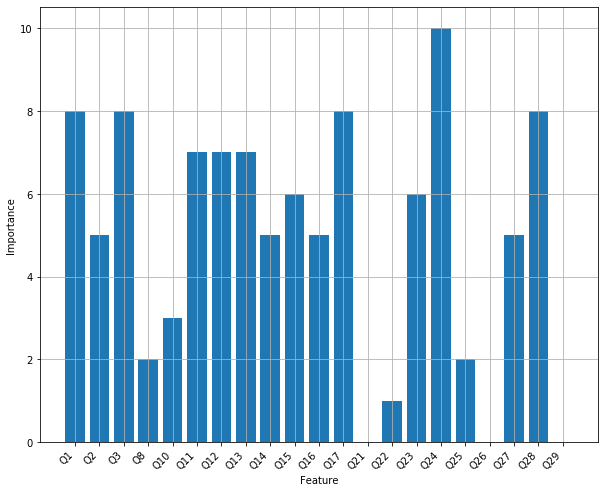}
  \label{fig13}
\end{figure}

LightGBM \cite{23} is another leaf-wise decision-tree based algorithm, and it works based on two novel techniques, Gradient-based One-Side Sampling (GOSS) and Exclusive Feature Bundling (EFB). The GOSS reduces the complexity by down sampling to remove examples with small gradients. The EFB improves the speed of the algorithm and reduces the complexity of it by down sampling features. In other words, the EFB bundles exclusive features into a single feature to reduce the complexity. LightGBM may outperform XGBoost, and it is often faster than XGBoost. One of the most important hyperparameters of LightGBM is number of trees. To find the most robust and accurate model, we tune number of trees between 2 and 40 while the number of leaves and the depth of each tree is left with no limit. Figure \ref{fig9} displays the accuracy scores of LightGBM as number of trees changes. It turns out that LightGBM with 20 trees performs better with an accuracy score of 94.51\%. Figure \ref{fig10} displays the feature importance scores of LightGBM with 20 trees.
We now control the number of leaves and the depth of trees to see whether we can improve the performance of LightGBM. Figure \ref{fig11} displays the accuracy scores of LightGBM with 20 trees as the depth of each tree and number of leaves change. Note that, we define number of leaves to be twice as large as the maximum depth of trees.
It turns out that LightGBM with 20 trees, where the depth of each tree is 4, obtains an accuracy of 95.6\%. Figure \ref{fig12} displays the feature importance scores of the LightGBM with 20 trees with depth 4.
Figures \ref{fig11} and \ref{fig12} indicate that LightGBM is significantly dependent on Questions 19 and 20 in the training process. To allow the model to explore further in details of other features, we remove questions 19 and 20 and retrain LightGBM on the remaining features. Figure \ref{fig13} displays the feature scores of LightGBM with 20 trees in absence of Questions 19 and 20.

\noindent \textbf{CatBoost:}

\begin{figure}[H]
  \centering
    \caption{Accuracy scores of CatBoost model as the number of trees changes between 4 and 19}    
  \includegraphics[width=\linewidth]{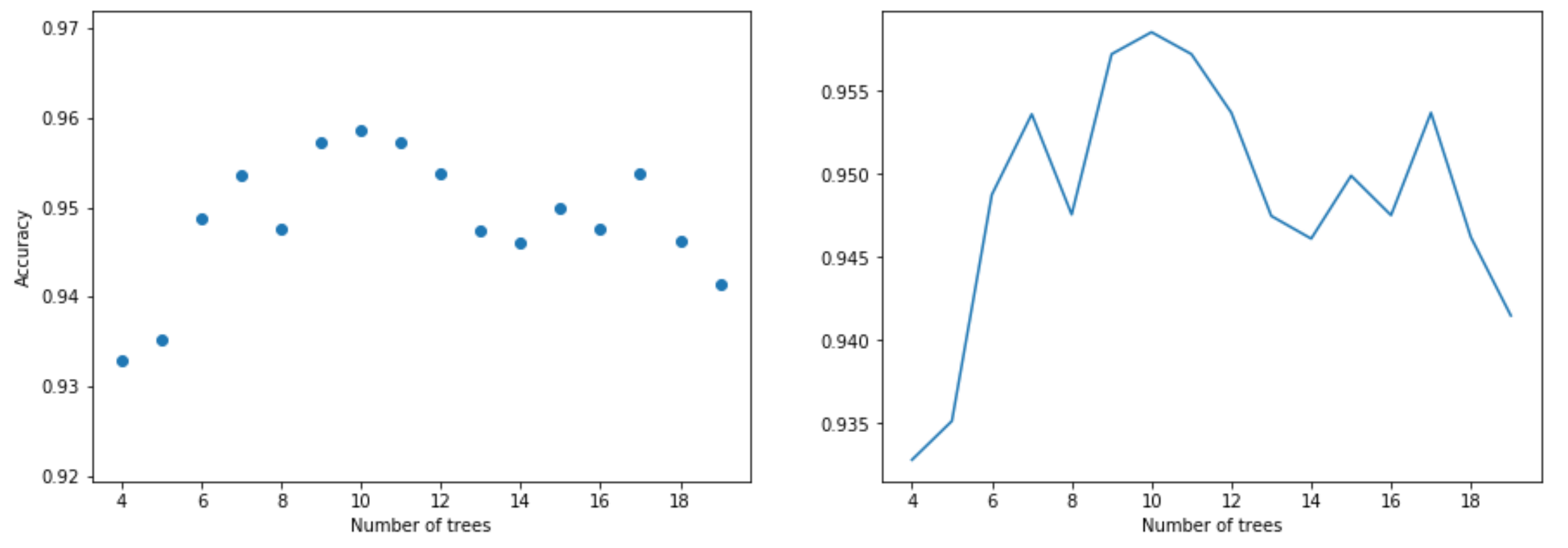}
  \label{fig14}
\end{figure}

\begin{figure}[H]
  \centering
    \caption{Feature scores of CatBoost with 11 trees and depth equals 6}    
  \includegraphics[width=.8\linewidth]{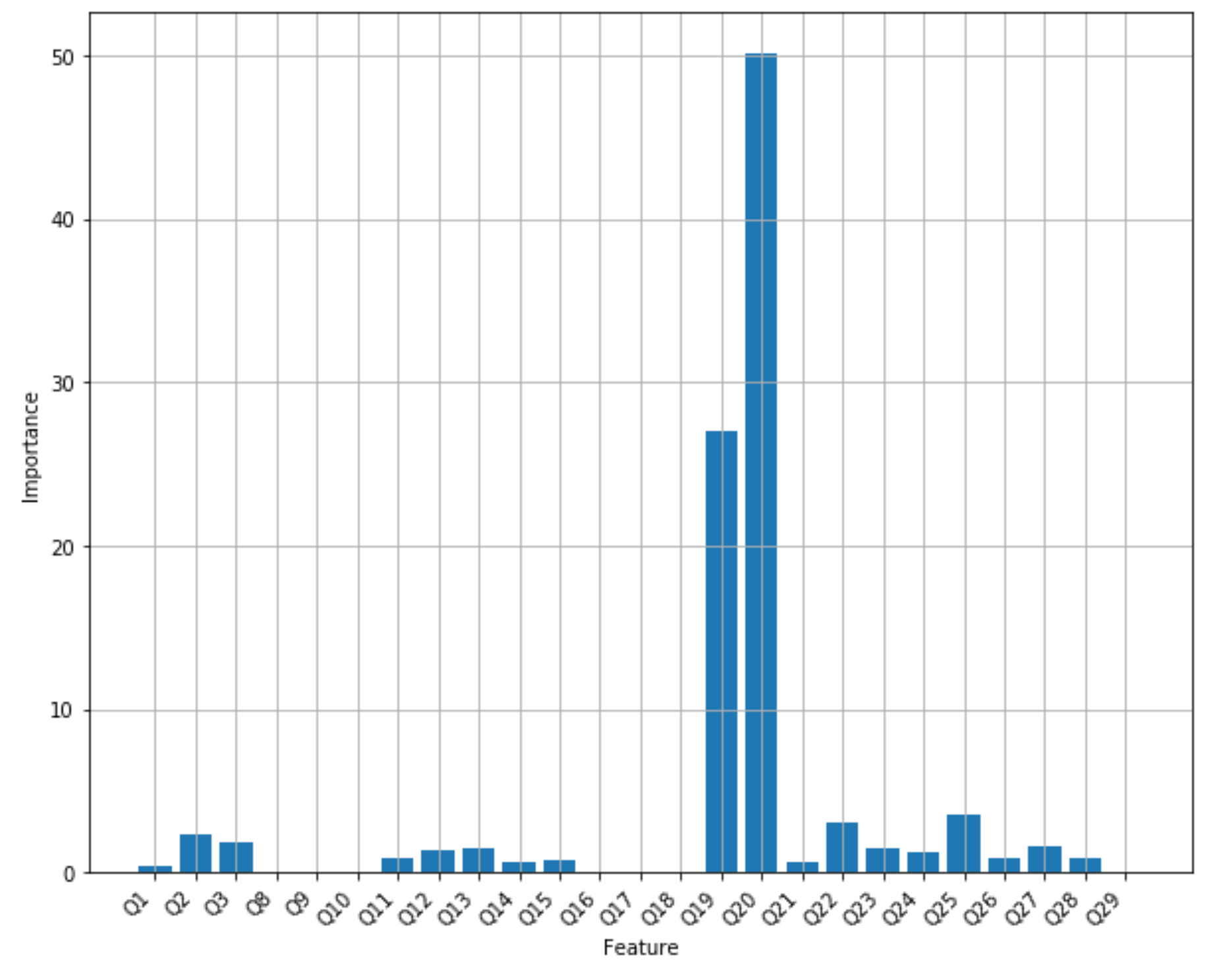}
  \label{fig15}
\end{figure}

\begin{figure}[H]
  \centering
    \caption{Feature scores of CatBoost with 11 trees and depth equals 6 in absence of Questions 19 and 20}    
  \includegraphics[width=.8\linewidth]{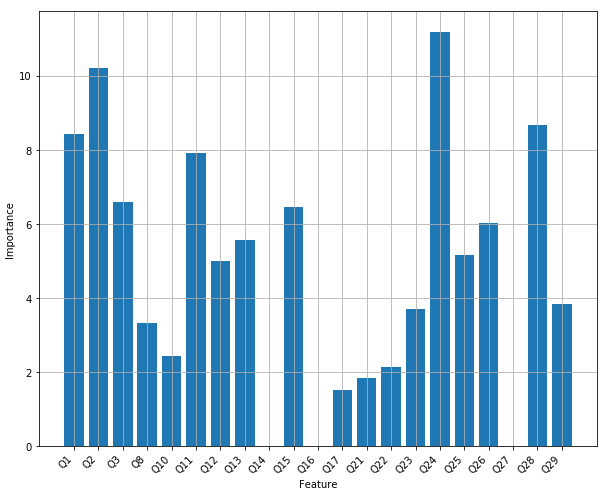}
  \label{fig16}
\end{figure}

CatBoost \cite{25} is another algorithm for gradient boosting on decision trees. It is a weighted sampling version of Stochastic Gradient Boosting that uses one-hot-encoding for categorical data. CatBoost utilizes two feature importance methods, the prediction-value-change and the loss-function-change. The prediction-value-change sorts features based on obtained prediction changes if a feature value changes. On the other hand, the loss-function-change sorts features based on the difference between loss value of the model with and without a feature. Number of trees and the depth of each tree are two important hyperparameters of CatBoost. Figure \ref{fig14} displays accuracy scores of CatBoost model while number of trees changes and the depth of each tree is fixed to be 6.  It turns out that CatBoost with 7 trees reaches the maximum accuracy of 95.4\%. Figure \ref{fig15} displays feature scores of CatBoost with 7 trees. 
Like XGBoost and LightGBM, CatBoost also makes its prediction mostly based on Questions 19 and 20. To find out the relationship between the target variable and other input variables, we remove Questions 19 and 20, and retrain CatBoost with 11 trees on the remaining features. Figure \ref{fig16} displays feature scores of CatBoost in absence of Questions 19 and 20. 
The most important features, or the top predictors, of the retrained CatBoost model are: 
\begin{itemize}
\item Question 24: ‘How many hours of COVID-related news or social media are you consuming on average per day?’’ 
\item Question 2: “What is your gender?’’
\item Question 28:” Has the amount of food you have been eating per day changed?’’
\item Question 1: “What is your age?’’
\end{itemize}

\begin{figure}[h]
  \centering
    \caption{Feature scores of SMOTE Random Forest with 100 trees and depth equals 8}    
  \includegraphics[width=.9\linewidth]{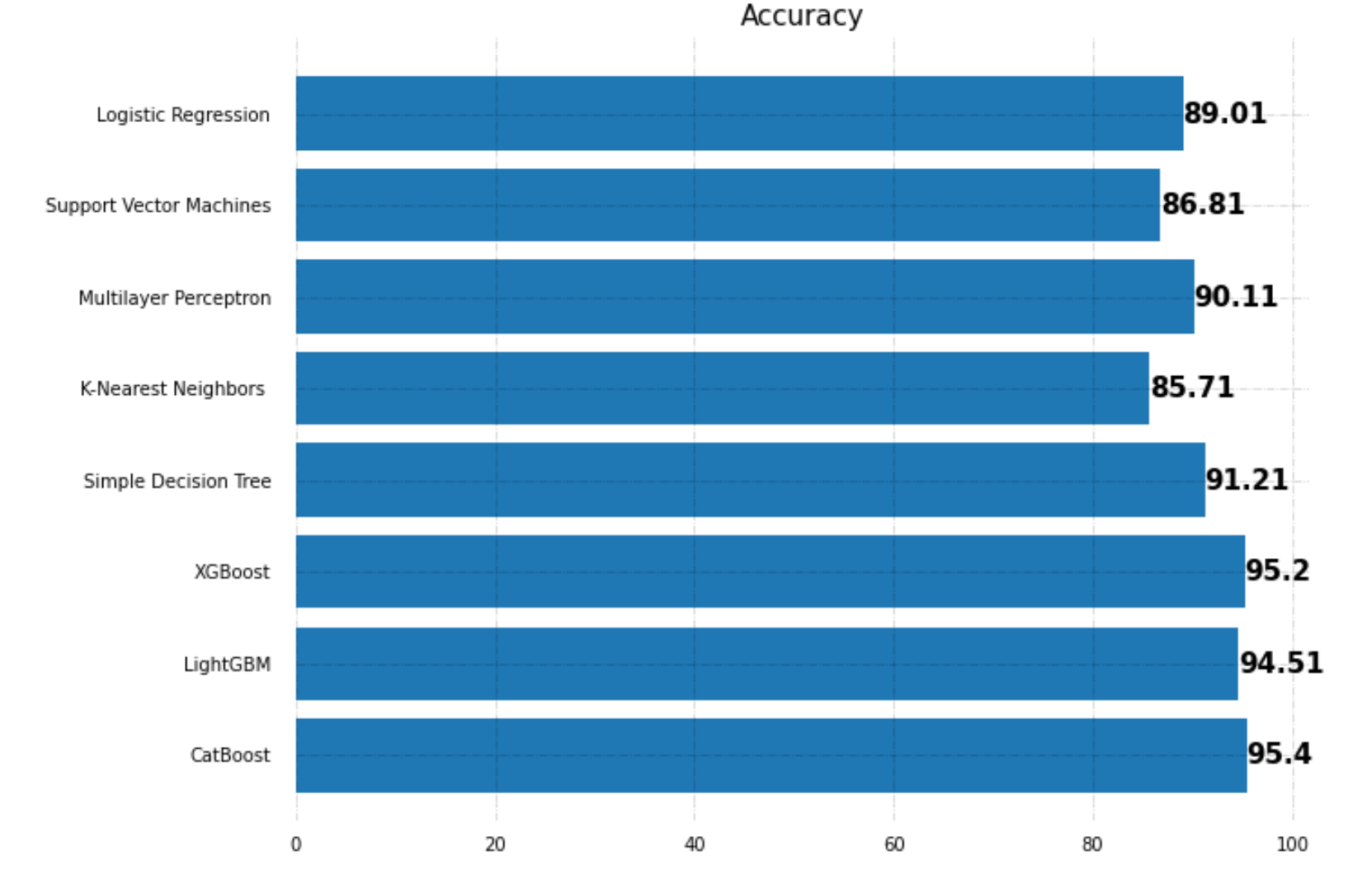}
  \label{fig1921}
\end{figure}

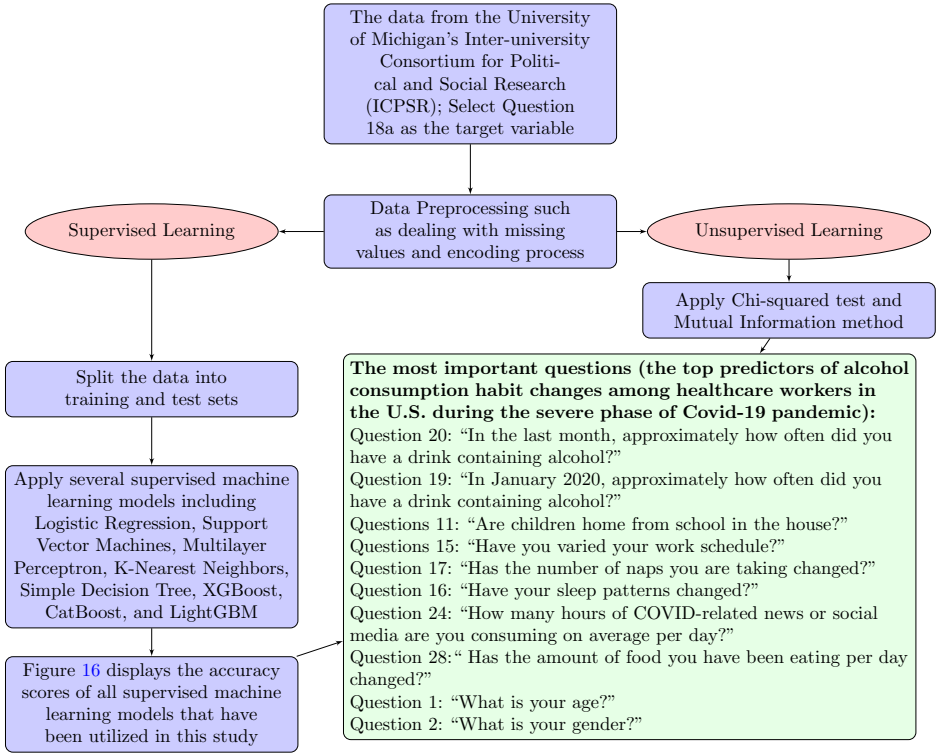
\begin{figure}[h]
\caption{A summary of the methology and the results}\label{Flowchart}
\begin{center}
\scalebox{.7}{%
\tikzstyle{decision} = [diamond, draw, fill=blue!20, 
    text width=4.5em, text badly centered, node distance=3cm, inner sep=0pt]
\tikzstyle{block} = [rectangle, draw, fill=blue!20, node distance=3cm,
    text width=15em, text centered, rounded corners, minimum height=3em]
\tikzstyle{blockw} = [rectangle, draw, fill=blue!20, node distance=1.5cm,
    text width=15em, text centered, rounded corners, minimum height=3em]
\tikzstyle{blockkk} = [rectangle, draw, fill=blue!20, node distance=6cm,
    text width=10em, text centered, rounded corners, minimum height=3em]
\tikzstyle{blockk} = [rectangle, draw, fill=green!10, node distance=9cm,
    text width=30em,  rounded corners, minimum height=10em]
\tikzstyle{line} = [draw, -latex']
\tikzstyle{cloud} = [draw, ellipse,fill=red!20, node distance=6cm,
    minimum height=3em]    
\begin{tikzpicture}[node distance = 3cm, auto]
    \node [block] (init) {The data from the University of Michigan's Inter-university Consortium for Political and Social Research (ICPSR); Select Question 18a as the target variable};
    \node [block, below of=init] (identify) {Data Preprocessing such as dealing with missing values and encoding process};
    \node [cloud, right of=identify] (expert) {Unsupervised Learning};
    \node [cloud, left of=identify] (system) {Supervised Learning};
    \node [block, below of=system] (identify2) {Split the data into training and test sets};

    \node [block, below of=identify2] (identify21) {Apply several supervised machine learning models including Logistic Regression, Support Vector Machines, Multilayer Perceptron, K-Nearest Neighbors, Simple Decision Tree, XGBoost, CatBoost, and LightGBM};
    \node [block, below of=identify21] (identify211) {Figure \ref{fig1921} displays the accuracy scores of all supervised machine learning models that have been utilized in this study};
    \node [blockw, below of=expert] (identify3) {Apply Chi-squared test and Mutual Information method};
 \node [blockk, right of=identify21] (Final) {\textbf{The most important questions (the top predictors of alcohol consumption habit changes among healthcare workers in the U.S. during the severe phase of Covid-19 pandemic):} \\
Question 20: “In the last month, approximately how often did you have a drink containing alcohol?’’ \\
Question 19: “In January 2020, approximately how often did you have a drink containing alcohol?’’\\
Questions 11: “Are children home from school in the house?’’\\
Questions 15: “Have you varied your work schedule?’’\\
Question 17: “Has the number of naps you are taking changed?’’\\
 Question 16: “Have your sleep patterns changed?’’\\
Question 24: “How many hours of COVID-related news or social media are you consuming on average per day?’’\\
Question 28:`` Has the amount of food you have been eating per day changed?’’ \\
Question 1: “What is your age?’’ \\
Question 2: “What is your gender?’’ };

    \path [line] (init) -- (identify);
    \path [line] (identify) --(expert);
    \path [line] (identify) -- (system);
    \path [line] (expert) -- (identify3);
    \path [line] (system) -- (identify2);
    \path [line] (identify2) -- (identify21);
    \path [line] (identify21) -- (identify211);
    \path [line] (identify211) --(Final);
    \path [line] (identify3) -- (Final);
\end{tikzpicture}
}
\end{center}
\end{figure}

Figure \ref{fig1921} displays the accuracy scores of all supervised machine learning models that discussed in this section. Top predictors for mental health analysis have been identified from different approaches. Among all the approaches, the random forest using SMOTE is the model that has identified the maximum top predictors. Figure \ref{Flowchart} summarizes the methology and the results in this section. 

\section{Discussion }\label{Section3}
In this section, we discuss the top predictors of alcohol consumption habit changes among healthcare workers in the United States that have obtained in Section \nameref{Section2} by several supervised and unsupervised machine learning models and methods. We also discuss some research articles with non machine learning approaches whose results agree with our results. 

Question 20: “In the last month, approximately how often did you have a drink containing alcohol?’’ is the most important feature for all supervised learning algorithms, Chi-Squared test and Mutual-Information method. Moreover, Question 19: “In January 2020, approximately how often did you have a drink containing alcohol?’’ is the second important feature in supervised learning algorithms. But, due to the nature of Question 18a, the relationship among Question 18a, Question 19 and Question 20 is clear.

One of the most important questions, or features, that appears in unsupervised learning methods as well as some supervised learning algorithms is Questions 11: “Are children home from school in the house?’’ It raises a question about the relationship between COVID-19 associated school closure and alcohol consumption habit changes among healthcare workers. There is a strong relationship between COVID-related school closure, parenting stress and alcohol consumption that requires the deep research of the topic. Based on our findings, which agree with the findings in \cite{35}, we make a conclusion about the relationship between parenting stress and alcohol consumption: the school closure associated with the COVID-19 pandemic can be associated with an increase in stress and anxiety among healthcare workers, which may lead to an increase in alcohol consumption as a coping mechanism. To alleviate hidden effects of COVID-19, e.g. parenting stress, on healthcare workers, some operational strategies might be used in schools to reduce the spread of COVID-19 and maintain safe operations in school during the COVID-19 pandemic. 

Both Chi-Squared test and Mutual-Information method as well as some supervised learning algorithms imply that Questions 15: “Have you varied your work schedule?’’ is not independent of alcohol consumption habit changes among healthcare workers. As the COVID-19 pandemic continues, healthcare workers are more in need to work shifts, and consequently healthcare workers who work shifts may consume more alcohol than dayworkers.  Our results support a study from The Ohio State University College of Nursing \cite{27} that reveals how the COVID-19 pandemic associated long shifts has impacted nurses working. In another study by Cooper et al. \cite{37}, it is shown that work stressors lead to increased distress, which in turn promotes problematic alcohol use. The findings in \cite{37} also reveals the relationship between work-related stress and an increase in alcohol consumption among healthcare professionals during the COVID-19 pandemic.

Next question that is selected by XGBoost and LightGBM is Question 17: “Has the number of naps you are taking changed?.’’ Moreover, Question 16: “Have your sleep patterns changed?’’ is selected by XGBoost as an important feature. One way that we can justify the relationship between alcohol consumption changes and sleep pattern changes is to give close and thoughtful attention to the findings of \cite{28}. According to a study titled Alcohol Alert by the National Institute on Alcohol Abuse and Alcoholism \cite{28}, alcohol consumption can change sleep patterns. One way that healthcare workers may choose to cope with the COVID-19 associated parenting stress and long shifts is by turning to alcohol, which leads to a change in sleep pattern. So, to reduce sleep problems associated with COVID-19 among healthcare workers, some psychology-based strategies may be used in hospitals to decrease work-related stress and the usage of alcohol in healthcare workers and keep them safe. 

LightGBM and CatBoost select Question 24: “How many hours of COVID-related news or social media are you consuming on average per day?’’ as one the most important features in their training process. Buchanan et al. \cite{29} examines the emotional consequences of exposure to COVID-related news. The findings of [29] can help us have a better understanding on the relationship between COVID-related news exposure and alcohol use changes. Due to their job, healthcare workers seek out COVID-19 information as a means of coping with challenging situations, and as a result they put themselves in stressful situations. Strainback et al. \cite{30} also explores COVID-related news effects on mental health. Their findings show that COVID-19 media consumption leads to psychological distress, which may cause an increase in alcohol consumption.  Thus, we can make a conjecture: the stress resulting from COVID-19 related news exposure may produce changes in drinking behavior. 

Question 28:`` Has the amount of food you have been eating per day changed?’’ is another important feature that is selected by LightGBM and CatBoost. An increase in alcohol consumption can cause changes in a healthcare worker diet. People who drink alcohol more frequently may choose less healthy food options that are high in fat and sugar \cite{31}. Moreover, each gram of pure alcohol has 29 kilojoules of energy \cite{32}, and when it is mixed with sugary drinks, it contains even more calories.  Since alcohol is absorbed directly in the bloodstream \cite{33} that can lead to immediate changes on the amount of food that people eat. 

Finally, Question 1: “What is your age?’’ and Question 2: “What is your gender?’’ are selected as important features by LightGBM and CatBoost. Lavretsky et al. \cite{34} examines stress-related changes associated with aging and sex differences. Their results can help us have a better understanding about the relationship between alcohol consumption changes due to COVID-related stress, aging and gender differences. Novais et al. \cite{36} also examines the effects of age and sex differences in the stress response. Their findings agree with the findings of \cite{34} and lead to the same result. Therefore, regarding the relationship between age and alcohol consumption changes, we may utilize our findings and the findings of \cite{38} to make a conclusion: the susceptibility to stress differs between young and old people, and may lead to different stress responses e.g., turning to alcohol to cope with stressful situations. Regarding the relationship between gender differences and alcohol consumption changes, we may give a close attention to the results of \cite{39} and make a conclusion: men and women tend to react differently with stress, both psychologically and biologically, and may lead to different stress responses. 

\section{Conclusions}\label{Section4}
In this paper, we have examined the effects of the COVID-19 pandemic on alcohol consumption habit changes among healthcare workers using a mental health survey data obtained from the University of Michigan Inter-University Consortium for Political and Social Research. We have utilized several supervised and unsupervised machine learning methods and models such as Decision Trees, Logistic Regression, Naive Bayes classifier, k-Nearest Neighbors, Support Vector Machines, Multilayer perceptron,  XGBoost, CatBoost, LightGBM, Chi-Squared Test and mutual information method to find out how the COVID-19 pandemic causes changes in alcohol use and stress.  Through the interpretation of many supervised and unsupervised methods applied to the dataset, we have concluded that some effects of the COVID-19 pandemic such as school closure, work schedule change and COVID-related news exposure may lead to an increase in alcohol use. In future work, we are interested in analyzing more data related to healthcare workers to examine the relationship between their mental health and other factors related or unrelated to COVID-19. We are also interested in implementing other optimization methods such as those that are introduced in \cite{40} and \cite{41} to see whether accuracy or speed of some models, which were utilized in this paper, can be improved.

\end{document}